\documentclass{article}
\pdfoutput=1

\usepackage{PRIMEarxiv}

\usepackage[utf8]{inputenc} 
\usepackage[T1]{fontenc}    
\usepackage{hyperref}       
\usepackage{url}            
\usepackage{booktabs}       
\usepackage{amsfonts}       
\usepackage{nicefrac}       
\usepackage{microtype}      
\usepackage{amsmath,amssymb,amsfonts}
\usepackage{lipsum}
\usepackage{fancyhdr}       
\usepackage{graphicx}       
\graphicspath{{media/}}     

\title{\textbf{Only Positive Cases: 5-fold High-order Attention Interaction Model for Skin Segmentation Derived Classification\footnotemark[3]}}

\author{
  1\textsuperscript{st} Renkai Wu\footnotemark[2] \\
  Shanghai University \\
  Shanghai, China\\
  \texttt{wurk@shu.edu.cn} \\
   \And
  2\textsuperscript{nd} Yinghao Liu\footnotemark[2] \\
  Ruijin Hospital, Shanghai Jiao Tong University School of Medicine \\
  Shanghai, China\\
  \texttt{18616528897@163.com} \\
  \AND
  3\textsuperscript{rd} Pengchen Liang\footnotemark[1] \\
  Shanghai University \\
  Shanghai, China\\
  \texttt{liangpengchen@shu.edu.cn} \\
  \And
  4\textsuperscript{th} Qing Chang\footnotemark[1] \\
  Ruijin Hospital, Shanghai Jiao Tong University School of Medicine \\
  Shanghai, China\\
  \texttt{robie0510@hotmail.com} \\
}

\begin{document}
\maketitle
\renewcommand{\thefootnote}{\fnsymbol{footnote}}
\footnotetext[1]{Corresponding authors.}
\footnotetext[2]{These authors contributed equally to this work.} 
\footnotetext[3]{This work has been submitted to the IEEE for possible publication. Copyright may be transferred without notice, after which this version may no longer be accessible.}

\begin{abstract}
Computer-aided diagnosis of skin diseases is an important tool. However, the interpretability of computer-aided diagnosis is currently poor. Dermatologists and patients cannot intuitively understand the learning and prediction process of neural networks, which will lead to a decrease in the credibility of computer-aided diagnosis. In addition, traditional methods need to be trained using negative samples in order to predict the presence or absence of a lesion, but medical data is often in short supply. In this paper, we propose a multiple high-order attention interaction model (MHA-UNet) for use in a highly explainable skin lesion segmentation task. MHA-UNet is able to obtain the presence or absence of a lesion by explainable reasoning without the need for training on negative samples. Specifically, we propose a high-order attention interaction mechanism that introduces squeeze attention to a higher level for feature attention. In addition, a multiple high-order attention interaction (MHAblock) module is proposed by combining the different features of different orders. For classifying the presence or absence of lesions, we conducted classification experiments on several publicly available datasets in the absence of negative samples, based on explainable reasoning about the interaction of 5 attention orders of MHAblock. The highest positive detection rate obtained from the experiments was 81.0$\%$ and the highest negative detection rate was 83.5$\%$. For segmentation experiments, comparison experiments of the proposed method with 13 medical segmentation models and external validation experiments with 8 state-of-the-art models in three public datasets and our clinical dataset demonstrate the state-of-the-art performance of our model. The code is available from https://github.com/wurenkai/MHA-UNet.
\end{abstract}


\section{Introduction}
The skin, as the largest organ of our body, assumes the function of protecting our health. However, due to long-term exposure to ultraviolet radiation, melanocytes in the epidermis produce melanin in large quantities. The most serious form of skin cancer is melanoma, which is formed by the accumulation of melanin. According to the World Health Organization \cite{al2021survey}, about 132,000 people worldwide are affected by skin cancer (melanoma) every year. And according to the American Cancer Society data, the mortality rate of malignant melanoma reaches 7.66$\%$ \cite{hasan2023survey}. In particular, if it is diagnosed at a late stage, the survival rate will drop to 25$\%$ within 5 years. In addition to this, there exist a number of non-melanoma pigmented skin diseases (blue nevus, nevus of Ota and spitz nevus). Although they do not lead to death, they often develop in the facial area, which can have a major impact on the patient's appearance. In particular, as the quality of life continues to improve, people are placing more emphasis on their appearance.

Traditional manual diagnostic methods are severely limited by the level of expertise of dermatologists. The accuracy of diagnosis is often unsatisfactory, especially for areas where medical resources are scarce. According to \cite{tran2005assessing}, the accuracy of traditional manual diagnosis is between 24$\%$ and 77$\%$. And due to the failure of the diagnosis of the disease, it will lead to the malignant melanoma to miss the best time of diagnosis. Currently, computer-assisted diagnosis can greatly improve the diagnosis rate. For remote areas, as long as there is a computer, the level of diagnosis can be greatly improved. Among them, segmentation tasks are an important part of computer-aided diagnosis. However, in the traditional segmentation task, without training on negative samples, disappointing results are often obtained when segmenting regions without lesions. And this will lead to diagnostic errors. However, medical data is scarce. Based on this problem, our work proposes a segmentation task that only requires training in the segmentation task with positive samples, which gives excellent segmentation results and classification results.

Recently, in natural image segmentation, researchers have proposed a model with higher-order spatial interaction, HorNet \cite{rao2022hornet}. A high-order spatial interaction mechanism is proposed in HorNet, which can extract feature information at a deeper level. The key module gnconv can combine the advantages of convolution and Transformers at the same time, and also has the advantages of efficient, extendable and Translation-equivariant. Compared with ordinary convolution, gnconv is able to combine the spatial location information of lesions with the spatial location information of neighboring regions. Versus Transformers, gnconv extends the traditional self-concerned second-order interaction to an arbitrary order. This greatly improves the performance of the model \cite{rao2022hornet}. However, researchers have not deeply analyzed the differences in extracting features at different orders. One of the main contributions of this paper is to analyze the features extracted by different orders and propose multiple high-order attentional interaction mechanisms.

In skin lesion segmentation, lesions usually have more interferences, including skin epidermal shedding, hair interference, low contrast and boundary blurring, etc. In this paper, we propose a multiple high-order attention interaction U-shaped model for skin lesion segmentation. Specifically, a squeeze attention mechanism is added to each of the traditional high-order interactions. This is able to introduce the squeeze attention mechanism to higher interactions, allowing feature extraction to receive further attention at a deeper level. In addition, we use multiple high-order hybrid interactions in each layer in the UNet architecture. Specifically, as shown in Fig.\ref{fig03}, we propose a MHAblock, which is capable of simultaneous 1, 2, 3, 4 and 5 order attention interactions. Based on this, we can determine the presence or absence of a lesion simply based on the explainable features of the 5 different orders in order to avoid subsequent wrong diagnosis. And none of the model training process has negative samples for training. Our contribution is as follows:

\begin{itemize}
\item A Multiple High-order Attention Interaction Module (MHAblcok) is proposed for explainable skin lesion segmentation, combining different levels of feature information obtained from multiple different order interactions, and finally voting to select the best feature information for output.

\item An explainable segmentation method based on MHAblock is proposed and classification results are derived by explainable inference without the need to learn from negative samples.

\item An explainable inference classification algorithm (EICA) is proposed. EICA is able to determine the presence or absence of a lesion through explainability and does not introduce additional memory.

\item A high-order attention interaction module (HAblcok) is proposed. HAblock introduces the squeeze attention mechanism to each order of interaction, allowing feature information to be further attended to at a deeper level.

\item Our multiple high-order attention interaction mechanism is combined with the UNet architecture to propose the Multiple High-order Attention Interaction U-Shape Model (MHA-UNet). In the skip-connection part, MHA-UNet is successfully combined with Spatial Attention Module (SAB) and Channel Attention Module (CAB) for multilevel and multiscale information fusion.

\end{itemize}

\section{Related work}
Skin lesion segmentation is one of the very important studies in medical image segmentation. Among all types of global diseases, skin diseases have a high incidence and occupy a very large portion of global diseases \cite{hay2014global}. Many researchers have proposed many excellent algorithms to improve the accuracy and speed of automatic segmentation. Att-UNet \cite{oktay2018attention} is a model based on the UNet architecture. Att-UNet focuses on the lesion region and suppresses the redundant feature information by adding an attention mechanism to the UNet. In \cite{chen2021transunet}, Transformers-based UNet model is proposed. It applies both Transformers and convolution for the extracted features. SCR-Net \cite{wu2021precise} proposes two key modules (refinement module and calibration module) to enhance the recognition of lesion features. MedT \cite{asadi2020multi} proposed a solution to the problem of Transformers-based network architectures that require a large amount of data samples, which does not require pre-training on a large number of data samples. MedT uses gated axial attention as the main module of the encoder and trains the model with a LoGo strategy. FAT-Net \cite{wu2022fat} uses the Transformers encoder in the UNet model architecture uses Transformers encoder for the encoder part. The decoder part uses Feature Adaptation Module for multilevel feature fusion. ATTENTION SWIN U-NET \cite{aghdam2023attention} is optimized and improved based on Swin U-Net \cite{cao2022swin}, where the cascade operation incorporates an attention mechanism to suppress unimportant information. In \cite{ruan2022malunet}, a lightweight UNet model architecture is proposed. Lightweighting is achieved by using the fusion of multi-level and multi-scale feature information, which makes it possible to achieve the same performance without requiring a higher number of channels. C$^2$SDG \cite{gu2023cddsa} proposes a feature comparison enhancement method. It takes the input image and extracts the lower features and performs style enhancement for contrast training. In \cite{azad2022transnorm}, an adaptive feature calibration method using an attention mechanism in the skip-connection part is proposed. META-Unet \cite{wu2023meta} proposes to use a combination of both high and low levels of transform attention. The use of two different levels of transformed attention improves the generalization ability of the model. MSA \cite{wu2023medical} is a general segmentation model based on the Segment Anything Model (SAM) \cite{kirillov2023segment} adapted for use in the medical domain. MSA is adapted by adding the Adapter module to the SAM to fit the segmentation in the medical domain.

In the above mentioned medical image segmentation models related to skin lesions, all are constructed with Transformers and convolution as the base module. Convolution cannot incorporate global information and Transformers training requires a large sample size. Even though many researchers currently use them in combination \cite{chen2021transunet, azad2022transnorm}, the respective shortcomings still cannot be eliminated. Currently, the higher-order spatial interaction mechanism proposed by researchers in natural scenes can solve the shortcomings of both of them well \cite{rao2022hornet}. In this paper, we take a further step in the high-order spatial interaction mechanism. We propose a high-order attention interaction mechanism and combine it with multiple high-order synthesized representations to form a multiple high-order attention interaction model (MSA-UNet). In the next subsection, we will elaborate our proposed multiple high-order attention interaction model.

\section{Method}
\subsection{Overall architecture}
\begin{figure}[!t]
\centering
\includegraphics[width=0.6\linewidth]{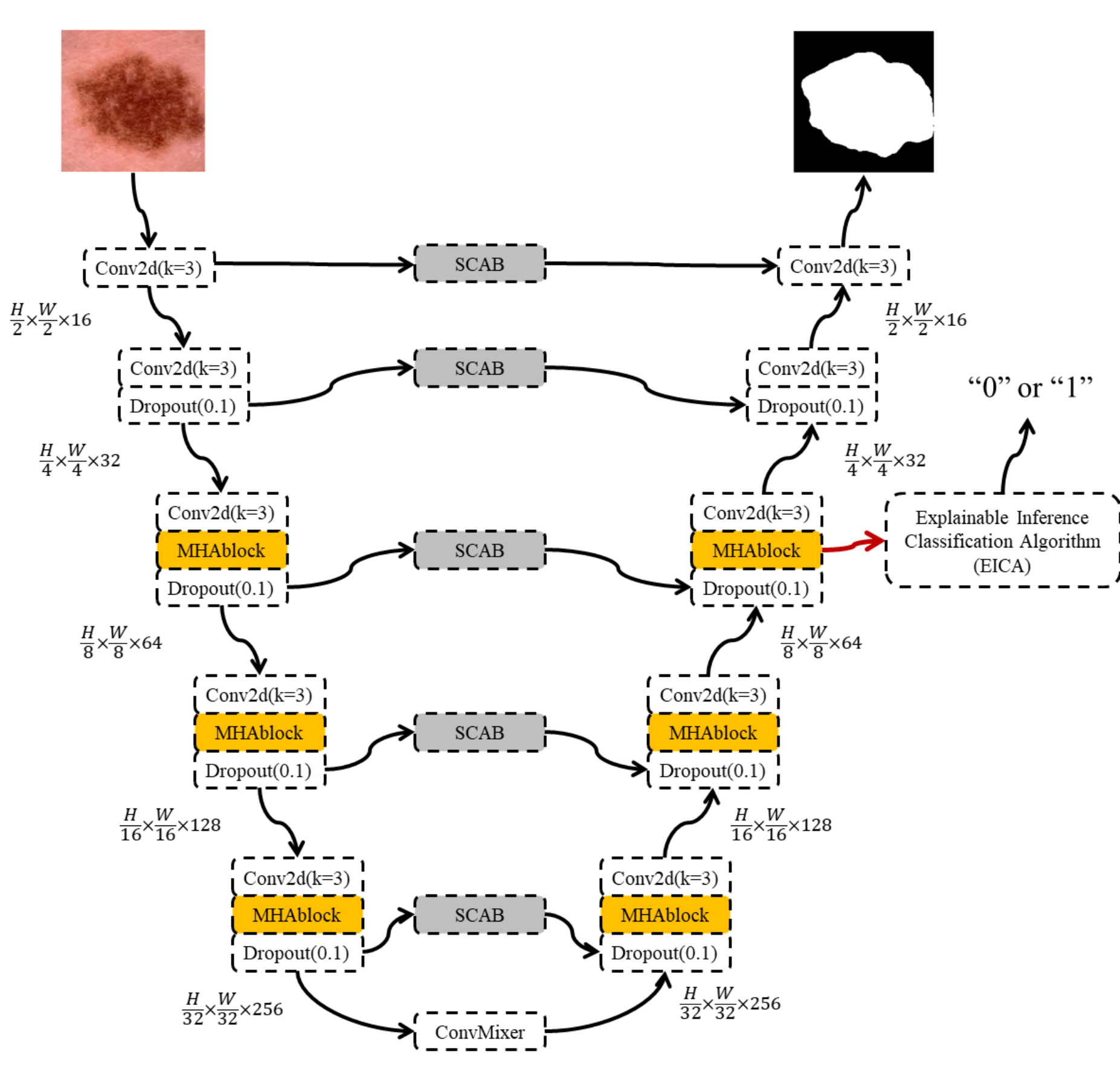}
\caption{The MHA-UNet model architecture proposed in this paper.}
\label{fig01}
\end{figure}

The overall architecture of the proposed Multiple High-order Attention Interaction Model (MSA-UNet) can be shown in Fig.\ref{fig01}. MSA-UNet adopts a UNet-like architecture consisting of an encoder, a skip-connection part, and a decoder, respectively. The MSA-UNet employs a 5-layer UNet architecture, which is carried out layer by layer to reduce the image size and increase the number of channels. The number of channels is set to [16,32,64,128,256]. Layer 1 uses a standard convolution (convolution kernel of 3) for feature extraction. Layer 2 uses the same standard convolution and a Dropout operation. From layer 2 to layer 5, a Dropout operation is used to prevent model overfitting. Layers 3 through 5 go through a standard convolution, Multiple High-Order Attention Interaction Block (MHAblock), and a Dropout operation, respectively. The ConvMixer operation is used after layer 5 to improve the generalization of the model. The decoder and encoder are kept symmetric, which helps in effective fusion of features. The skip-connection part uses Channel Attention Module (CAB) and Spatial Attention Module (SAB) for multilevel and multiscale fusion. The use of channel attention module and spatial attention module accelerates the model fitting.

The key module of the Multiple High-order Attention Interaction model (MSA-UNet) is the Multiple High-order Attention Interaction module (MHAblock). The key module of the Multiple High-order Attention Interaction Module (MHAblock) is the High-order Attention Interaction Module (HA-blcok). In the next subsections, we will focus on these modules.

\subsection{High-order Attention Interaction Module}
\begin{figure}[!t]
\centering
\includegraphics[width=0.6\linewidth]{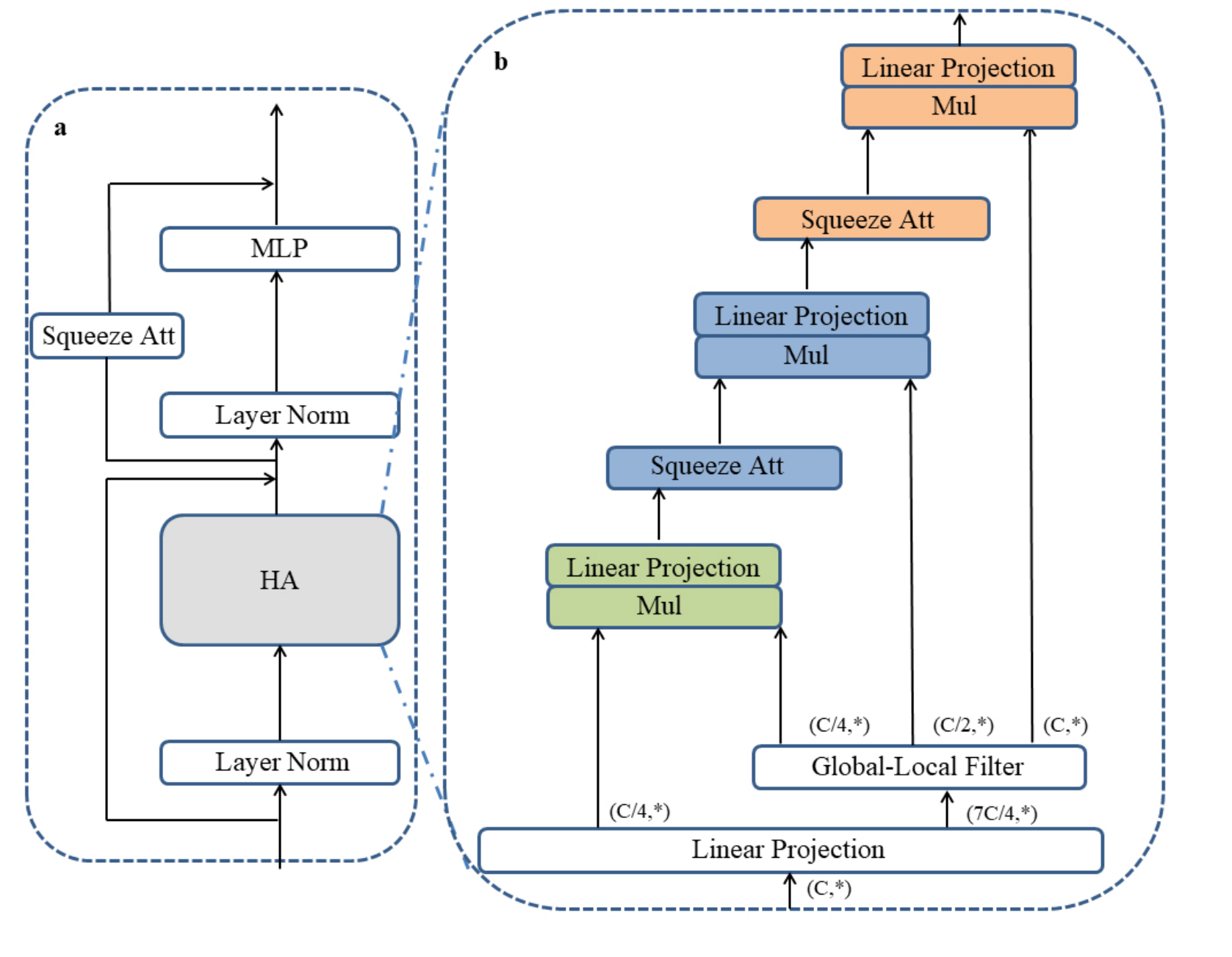}
\caption{(a)The HA-block structure proposed in this paper. (b)The HA structure proposed in this paper. Different colored sections represent different orders of attention interactions.}
\label{fig02}
\end{figure}

Traditional high-order spatial interactions \cite{rao2022hornet} lack focus on features. Specifically, as the interaction order increases with increasing order, the lesion features are gradually lost due to the continuous fusion with the global information. To address this problem, as shown in Fig.\ref{fig02}, we propose the Higher Order Attention Interaction Block (HA-block). The proposed HA-block employs a structure similar to Transformers. The self-attention layer inside Transformers is only capable of second-order interactions. HA-block replaces the self-attention layer with a high-order attention interaction layer (HA). HA mainly consists of squeeze attention, linear projection layer, and global-local filter. HA enables the manipulation of long-term, high-order attention, and high-order spatial interactions.

\textbf{2-order attention interactions} In order to detail the high-order attention interaction mechanism, we start from the 2-order attention interaction operation. The squeeze attention operation was not used for the 1-order interaction. This is due to the requirement of the squeeze attention operation to act on features that are fused with the global-local filter. Assuming $x \in \mathbb{R}^{H W \times C}$ is an input feature, we can express the 2-order attention interaction operation by the following equation:
\begin{equation}
\left[X_{0}^{H W \times C}, Y_{0}^{H W \times C_0}, Y_{1}^{H W \times C_1}\right]=Pro_{i n}(x) \in \mathbb{R}^{H W \times 2 C}
\end{equation}
\begin{equation}
Out_1=Pro\left[X_{0} * G L F\left(Y_{0}\right)\right] \in \mathbb{R}^{H W \times C}
\end{equation}
\begin{equation}
Out_2=Pro_{out}\left[S A\left(Out_1\right) * G L F\left(Y_{1}\right)\right] \in \mathbb{R}^{H W \times C}
\end{equation}
\noindent
where $Pro_{i n}$ and $Pro_{out}$ are the input and output linear projection layers respectively, the input features are doubled in the number of channels and two features ($X_{0}$ and $Y_{0}$) are formed after passing through $Pro_{i n}$. $SA$ is the squeeze attention module and $GLF$ is the global-local filter. After multiplication and output linear projection layer operation, the 1-order attention interaction operation of two features $X_{0}$ and $Y_{0}$ can be realized.

\textbf{High-order attention interactions} High-order attention interactions can be generalized by 1-order attention interactions. Suppose $x \in \mathbb{R}^{H W \times C}$ is an input feature. The input features will output an $X_0$ and a set of $Y$ features after passing through the linear projection layer with a total number of channels of 2C. The number of channels of each order can be obtained from $C_{k}=\frac{C}{2^{n-k-1}}, 0 \leq k \leq n-1$. Specifically, the high-order attention interaction operation may be expressed by the following equation:
\begin{equation}
\left[X_{0}^{H W \times C_{0}}, Y_{0}^{H W \times C_{0}}, \cdots Y_{n-1}^{H W \times C_{n-1}}\right]=Pro_{i n}(x) 
\end{equation}
\begin{equation}
X_{k+1}=Pro\left[SA\left(X_{k}\right) * G L F\left(Y_{k}\right)\right] / \alpha
\end{equation}
\noindent
where $\alpha$ is a stabilization factor that stabilizes the training of the model. After each interaction, $k$ is increased by 1, and each order process performs squeeze attention and $GLF$ operations. High-order attention interactions can be performed through the above operations.

\begin{figure*}[!t]
\centering
\includegraphics[width=\linewidth]{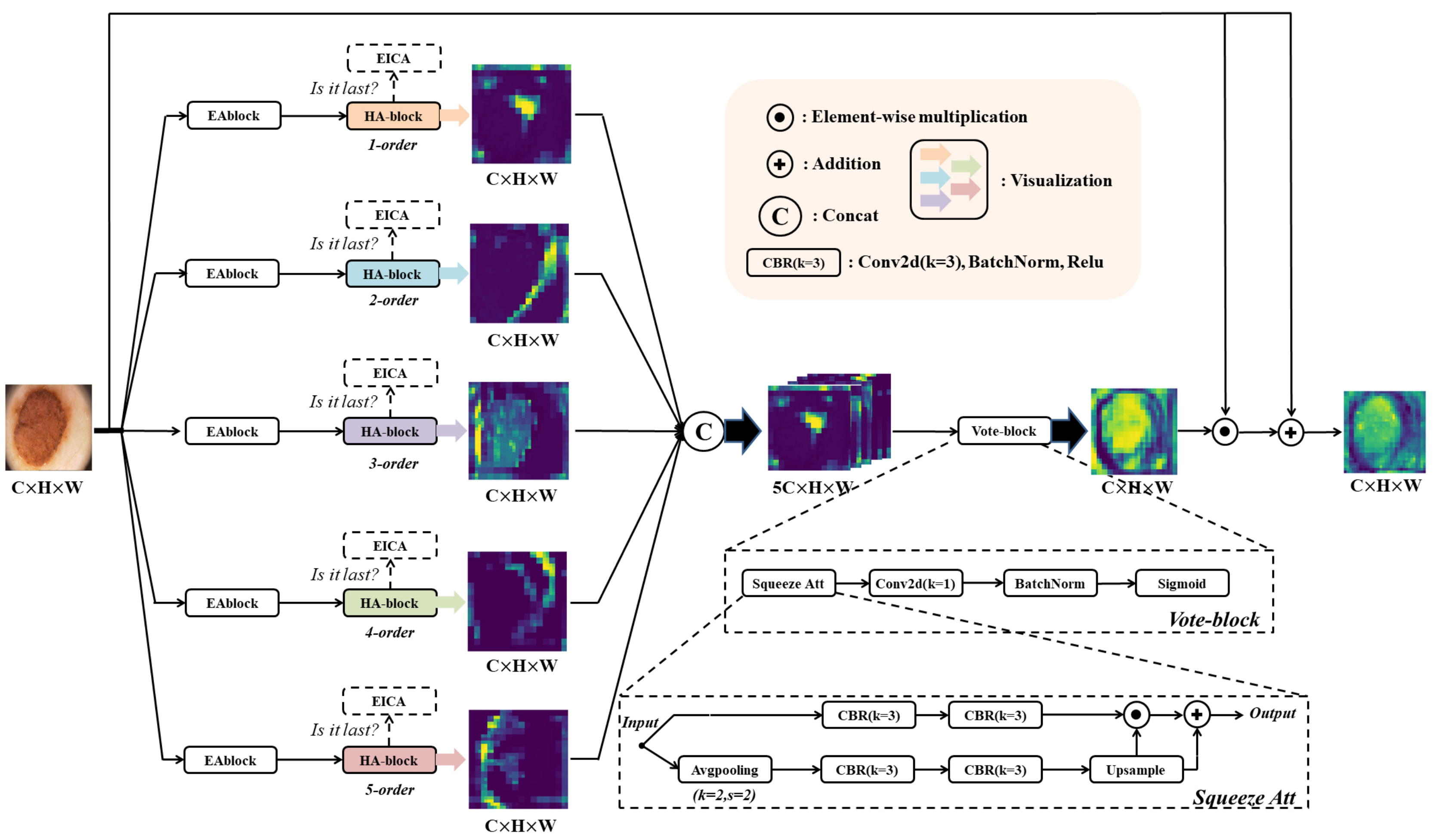}
\caption{The MHAblock structure proposed in this paper.}
\label{fig03}
\end{figure*}

\textbf{Squeeze Attention} In \cite{zhong2020squeeze}, researchers designed squeeze attention module for segmentation task based on the segmentation task characteristics. In segmentation task, convolution extracts features by localizing near each pixel. While in the global image attention process, the focus objects in the feature map can be activated by operations such as global pooling, convolution and upsampling. SA is able to select the focus feature objects by weighting both globally and locally, which is extremely suitable for the segmentation task. Therefore, we propose for the first time to introduce the squeeze attention mechanism into higher-level interactions. The feature information will be learned more profoundly through the higher order level of squeeze and attention operations. Specifically, as in Fig.\ref{fig03} there is a structural diagram of squeeze attention. The specific operation can be expressed from the following equation:
\begin{equation}
\hat{O}_{att}=Conv_{att}\left[A P(x) ; \theta_{att}, \Omega_{att}\right]
\end{equation}
\begin{equation}
O_{att}=Up\left(\hat{O}_{a t t}\right)
\end{equation}
\begin{equation}
O=O_{att} * X_{res}+O_{att}
\end{equation}
\noindent
where $Conv_{att}$ is the attention convolution operation, parameterized by the attention factor $\theta_{att}$ and the convolutional layer structure $\Omega_{att}$. $AP$ is the average pooling layer, $Up$ is the upsampling operation, and $X_{res}$ is the output of the main convolutional channel.

\textbf{Global-Local Filter} Global-Local Filter (GLF) was proposed in \cite{rao2022hornet}. Earlier, the global filter (GF) was proposed in \cite{rao2021global}. The GF is added to the learning of frequency domain features by means of learnable parameters. The global-local filter was proposed to improve the global filter. The GLF extracts local features by feeding half the number of channels to a standard convolution (kernel of 3). The locally extracted features are then concatenated with the features learned in the global frequency domain. The frequency domain features are learned using mainly 2D Discrete Fourier Transform (2D FFT) and 2D Discrete Inverse Fourier Transform (2D IFFT). The frequency domain feature learning can learn spatial location information at the frequency domain level. And the inclusion of local feature learning can reduce the loss of lesion feature information when the model is increasing in the interaction order. The choice of global filtering in the frequency domain is due to the fact that filters learned in the frequency domain are more clearly expressed than in the spatial domain [HorNet]. The process of learning in the frequency domain can be expressed by the following: 
\begin{equation}
\mathcal{F}[f]:\mathbf{\mathrm{\mathbf{\mathrm{} } } } F(u, v)=\frac{1}{H W} \sum_{x=0}^{H-1} \sum_{y=0}^{W-1} f(x, y) e^{-/ 2 \pi(k x / H+v y / W)}
\end{equation}
\begin{equation}
\mathcal{F}^{-1}[F]:f(x, y)=\sum_{u=0}^{H-1} \sum_{v=0}^{W-1} F(u, v) e^{j 2 \pi(u x / H+v y / W)}
\end{equation}
\begin{equation}
F_{out}=\mathcal{F}^{-1}[\mathcal{V}*\mathcal{F}[f(x,y)]]
\end{equation}
\noindent
where $HW$ denotes the image size $H \times W$, $f(x,y)$ is the input two-dimensional image information, $x$ and $y$ denote the spatial domain image variables, $u$ and $v$ denote the frequency variables in the frequency domain, $\mathcal{F}$ denotes the two-dimensional Discrete Fourier Transform (DFT), $\mathcal{F}^{-1}$ denotes the two-dimensional In-verse Discrete Fourier Transform (IDFT), and $\mathcal{V}$ denotes the learnable filter. In fact, the Fast Fourier Transform (FFT) and the Inverse Fast Fourier Transform (IFFT) are mainly used in computers. The FFT is similar to the DFT, but the amount of operations in the FFT is reduced dramatically.

The high-order attention interaction mechanism formed by combining the global-local filter with squeeze attention is one of the important ideas we propose. Squeeze attention incorporates global properties after frequency-domain filtering for feature attention and introduces squeeze attention to higher-order attentional interactions via higher-order properties. Express the higher order attention interaction operation from the aspect of frequency domain filtering as:
\begin{equation}
\varGamma_o^k=O^k*Cat[Conv(F_{out}^k),L]
\end{equation}
\begin{equation}
X_{k+1}=Pro(\varGamma_o^k)/ \alpha
\end{equation}
\noindent
where $\varGamma_o^k$ is the feature of the $k$-order attention interaction, $Cat$ is the Concat operation, $Conv$ is the standard convolution (convolution kernel of 1) operation, and $L$ is the local feature learning in the global-local filter (standard convolution with convolution kernel of 3).

\subsection{Multiple High-order Attention Interaction}
Skin lesion characteristics are diverse and highly intrusive. Conventional higher order interactions simply use only individual order interactions. We found that different orders extract features with different levels of information of the features. As shown in Fig.\ref{fig03} is our proposed Multiple High-order Attention Interaction Block (MHAblock).

We visualized graphs of the results of different orders of attention interactions for MHAblock. The 1-order, 2-order, 3-order, 4-order and 5-order extract features with different levels of information respectively. 1-order extracts features with information about the location of the upper part of the lesion. 2-order extracts features with information about the boundary features on the lower right side of the lesion. 3-order extracts features with information about the overall outline of the lesion. 4-order extracts features with information about the boundary features on the upper right side of the lesion. 5-order extracts features with information about the boundary features on the left side of the lesion. It is not a coincidence that although we visualized only one sample of skin lesions, we found that all samples were the result of such feature extraction.

Traditional high-order interaction methods use only one order extraction. We propose a multiple high-order interaction approach by further analyzing the high-order interaction properties in depth. At the same time, the multiple high-order attention interaction module is proposed in combination with the proposed high-order attention interaction module. As shown in Fig.\ref{fig03}, we combine the interaction results of 5 orders for Concat connected, and the number of channels becomes 5 times. The most suitable feature information is filtered by a Vote-block, and the number of channels returns to the original. Specifically, the concrete composition of the multiple higher-order attention interaction module can be expressed by the following equation:
\begin{equation}
X=E A(x)
\end{equation}
\begin{equation}
\left.\begin{array}{l}
x_{1} \\
x_{2} \\
x_{3} \\
x_{4} \\
x_{5}
\end{array}\right\}=\left\{\begin{array}{l}
H A_{1}(X) \\
H A_{2}(X) \\
H A_{3}(X) \\
H A_{4}(X) \\
H A_{5}(X)
\end{array}\right.
\end{equation}
\begin{equation}
C_{0}=Concat\left(x_{1}, x_{2}, x_{3}, x_{4}, x_{5}\right).
\end{equation}
\begin{equation}
O u t=x * V b \left(C_{0}\right)+x
\end{equation}
\noindent
where $EA$ is Inverted External Attention Block (IEAB) \cite{ruan2022malunet}, and IEAB can improve the model generalization ability. $HA_1$, $HA_2$, $HA_3$, $HA_4$, and $HA_5$ denote the 1-order, 2-order, 3-order, 4-order and 5-order attention interaction operations, respectively. $Vb$ is Vote-block, and the specific Vote-block can be expressed by the following equation:
\begin{equation}
A=SA(x)
\end{equation}
\begin{equation}
Out=Sig\{B N[Conv(x)]\}
\end{equation}
\noindent
where $SA$ is the squeeze attention module, $Conv$ is the standard convolution with a convolution kernel of 1, $BN$ is the batch normalization operation, and $Sig$ is the Sigmoid activation function. 

\begin{figure}[!t]
\centering
\includegraphics[width=0.6\linewidth]{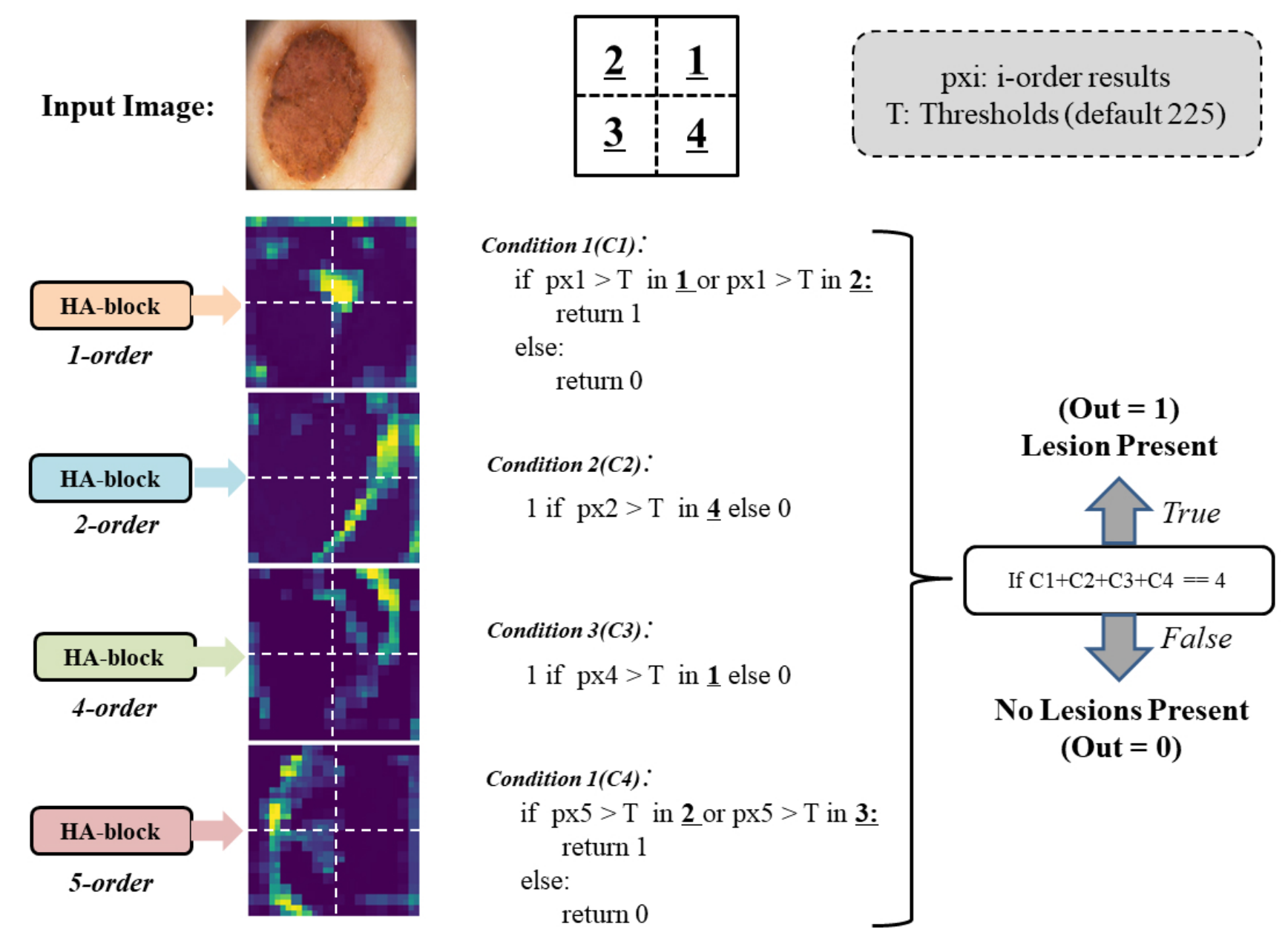}
\caption{Schematic diagram of the Explainable Inference Classification Algorithm (EICA) proposed in this paper.}
\label{fig04}
\end{figure}

\subsection{Explainable Inference Classification Algorithm (EICA) Design}
As known from Fig.\ref{fig04}, the HA-blocks of different orders of attention interactions have very distinctive features. For this reason, we design the Explanatory Inference Classification Algorithm (EICA) based on this strong feature. EICA is capable of determining the presence or absence of a lesion without the involvement of negative samples by the interpretability of attention interactions of different orders. EICA consists of strongly explainable mathematical formulas as shown in Fig.\ref{fig04}. EICA requires no training, is fast, and has a very low memory usage, and accessing the segmentation model takes up almost no additional memory.

The specific algorithm design is based on the characteristics of attention interactions of different orders. The image is divided into four quadrants, and the decision is made according to the learning prediction characteristics of 1-order, 2-order, 4-order and 5-order corresponding to the position where the feature information appears. As shown in Fig.\ref{fig04}, only if all four conditions are satisfied, we can determine it as the lesion exists (output is 1), otherwise output lesion does not exist (output is 0).

The proposed EICA has been integrated into the segmentation model to achieve end-to-end output classification and segmentation results under zero-negative sample training, and the classification task occupies almost no additional memory.

\section{Experiments}
\subsection{Datasets}
We used three public data on skin lesions (ISIC2017 \cite{codella2018skin}, ISIC2018 \cite{codella2019skin, tschandl2018ham10000} and PH$^2$ \cite{mendoncca2013ph}) and our clinical dataset. In particular, PH$^2$ and our clinical dataset are used as external validation to explore the generalization ability of the model due to their small sample size. The weights used for external validation were obtained after ISIC2017 training.

\begin{figure*}[!t]
\centering
\includegraphics[width=\linewidth]{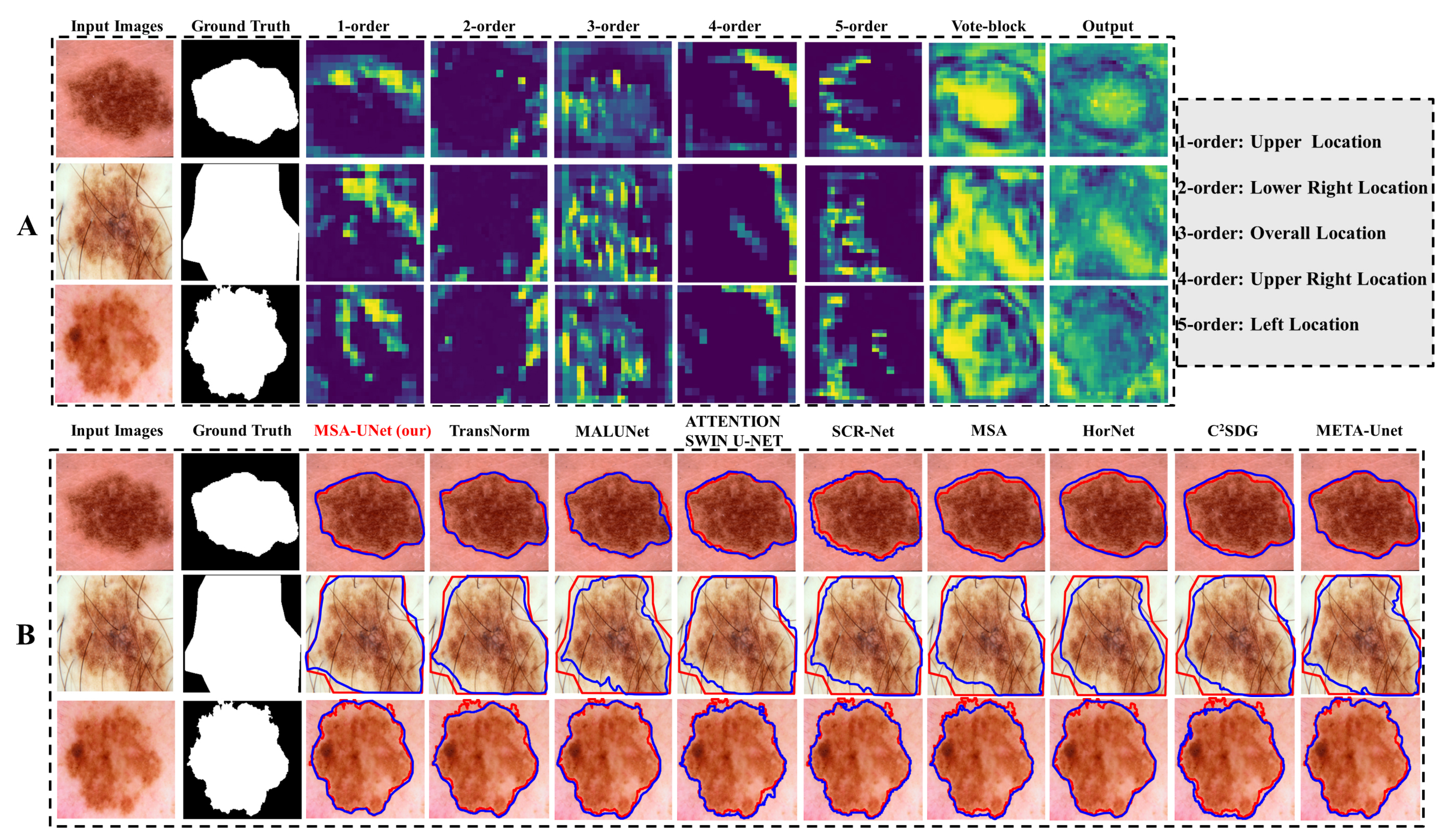}
\caption{(A)Visualization of the results of the individual modules in the last MHAblock of the decoder. (B)Visualization of segmentation results from multiple medical segmentation models and our model in ISIC2017 and ISIC2018. Red contours indicate ground truth and blue contours indicate model predicted segmentation lines.}
\label{fig05}
\end{figure*}

ISIC2017 and ISIC2018 acquire 2000 images and 2594 images, and the corresponding labels with segmentation masks, respectively. According to the settings of \cite{azad2022transnorm}, for the ISIC2017 dataset we used 1250 images for training, 150 images for validation and 600 images for testing. For the ISIC2018 dataset we used 1815 images for training set, 259 images for validation set and 520 images for testing set. The specific preprocessing was done as per \cite{alom2018recurrent} and the images were all resized to 256$\times$256 pixels.

The PH$^2$ dataset has a total of 200 images, and labels with segmentation masks. The image size is 768$\times$560 pixels. Due to the small number of 200 images, we use them as external validation. The specific preprocessing follows \cite{alom2018recurrent} and the images are all resized to 256$\times$256 pixels.

Our clinically acquired dataset was obtained from Ruijin Hospital, Shanghai Jiaotong University School of Medicine for 39 cases of skin lesions. Specifically, the clinical data were acquired by PhotoMax Pro Analyzer system using a magnification of 30$\times$. Three dermatologists with extensive experience performed the annotation of the true values. Due to the small sample size of the clinical data, we used it as an external validation. Again, all images were resized to 256$\times$256 pixels.

NormalSkin\footnote{https://universe.roboflow.com/janitha-prathapa/normalskin} dataset and a publicly available dataset inside Kaggle\footnote{https://www.kaggle.com/datasets/ahdasdwdasd/our-normal-skin/data} (denoted as Kaggle95 below) are both normal human skin images. They are mainly used to validate the negative detection rate of explainable inference classification trained without negative samples. After removing data with large viewpoints from both datasets, 200 and 95 normal skin images were obtained, respectively. Again, all images were resized to 256$\times$256 pixels.

\subsection{Implementation details}
All experiments are implemented on a single NVIDIA GeForce RTX 4080 Laptop GPU with 12GB of memory. And the experiments were realized based on Python 3.8 and Pytorch 1.12.0. For the training data, we used data augmentation (random rotation, vertical flip and horizontal flip) operations. BceDice \cite{ruan2022malunet} was used for the loss function. The training epoch was set to 250 and the batch size was 8. The optimizer used AdamW \cite{loshchilov2017decoupled}. The learning rate was minimized to 0.00001 with an initial value of 0.001 and a cosine annealed learning rate scheduler was used.

\subsection{Metrics}
Dice Score Coefficient (DSC), sensitivity (SE), specificity (SP) and accuracy (ACC) are the most commonly used evaluation criteria in the field of medical image segmentation. DSC is mainly used for evaluating the similarity between predicted and true values. SE mainly evaluates the percentage of true positives (TP) among true positives (TP) and false negatives (FN). SP mainly evaluates the measure of the percentage of false positives (FP) among percentage of true negatives (TN) and false positives (FP). ACC is used to evaluate the proportion occupied by true positives (TP) and true negatives (TN). In addition, the positive detection rate (PDR) and negative detection rate (NDR) in the classification task are consistent with the definitions of sensitivity (SE) and specificity (SP). It should be noted that PDR and NDR are in the form of percentages.
\begin{equation}
D S C=\frac{2\mathrm{TP}}{2\mathrm{TP}+\mathrm{FP}+\mathrm{FN}}
\end{equation}
\begin{equation}
A C C=\frac{\mathrm{TP}+\mathrm{TN}}{\mathrm{TP}+\mathrm{TN}+\mathrm{FP}+\mathrm{FN}}
\end{equation}
\begin{equation}
Sensitivity/ PDR =\frac{\mathrm{TP}}{\mathrm{TP}+\mathrm{FN}}
\end{equation}
\begin{equation}
Specificity/ NDR =\frac{\mathrm{TN}}{\mathrm{TN}+\mathrm{FP}}
\end{equation}
\noindent
where TP denotes true positive, TN denotes true negative, FP denotes false positive and FN denotes false negative.

\subsection{Comparison results}
In order to fully confirm the validity of our proposed method, we compared our experimental results with 13 medical segmentation models. In addition, we conducted external validation experiments with 8 of the most advanced models available.

Table \ref{tab1} and \ref{tab2} show the experimental results with 13 medical segmentation models on ISIC2017 and ISIC2018 datasets. We conclude from the table that the DSC value of the proposed MHA-UNet is 12.32$\%$ and 6.20$\%$ higher than the traditional UNet model on both data. The DSC values are 2.12$\%$ and 2.78$\%$ higher when compared with the currently popular Segment Anything Model in the medical field (MSA). In particular, the DSC values are also significantly higher when comparing with the traditional High order interaction model (HorNet) with a single order, respectively. Fig.\ref{fig05} shows the visualization result graph, and Fig.\ref{fig05}(A) shows the middle layer of the proposed MHAblock specific segmentation. We can see the model segmentation process very soberly, what kind of work is carried out in each order. This presents a powerful and explainable analysis for skin lesion segmentation. We fused multiple high-order attention interactions to get the best performance results with very clear interpretability, which is a difficult effect to achieve with other current models.

\begin{table}[!t]
\centering
\caption{Performance comparison with 13 medical segmentation models on ISIC 2017 dataset.}
\resizebox{0.75\linewidth}{!}{
\begin{tabular}{ccccc}
\hline
\textbf{Methods}     & \textbf{DSC}    & \textbf{SE}     & \textbf{SP}     & \textbf{ACC}    \\ \hline
U-Net \cite{ronneberger2015u}                & 0.8159          & 0.8172          & 0.9680          & 0.9164          \\
Att U-Net \cite{oktay2018attention}           & 0.8082          & 0.7998          & 0.9776          & 0.9145          \\
TransUNet \cite{chen2021transunet}           & 0.8123          & 0.8263          & 0.9577          & 0.9207          \\
MedT \cite{asadi2020multi}           & 0.8037          & 0.8064          & 0.9546          & 0.9090          \\
FAT-Net \cite{wu2022fat}              & 0.8500          & 0.8392          & 0.9725          & 0.9326          \\
TransNorm \cite{azad2022transnorm}            & 0.8933          & 0.8532          & 0.9859          & 0.9582          \\
MALUNet \cite{ruan2022malunet}              & 0.8896          & 0.8824          & 0.9762          & 0.9583          \\
ATTENTION SWIN U-NET \cite{aghdam2023attention} & 0.8859          & 0.8492          & 0.9847          & 0.9591          \\
SCR-Net \cite{wu2021precise}              & 0.8898          & 0.8497          & 0.9853          & 0.9588          \\
MSA \cite{wu2023medical}                  & 0.8974          & \textbf{0.9200} & 0.9824          & 0.9604          \\
HorNet \cite{rao2022hornet}               & 0.9063          & 0.9151          & 0.9746          & 0.9630          \\
C$^2$SDG \cite{gu2023cddsa}               & 0.8938          & 0.8859          & 0.9765          & 0.9588          \\
META-Unet \cite{wu2023meta}           & 0.9068          & 0.8801          & 0.9836          & 0.9639          \\
\textbf{MHA-UNet(Our)}    & \textbf{0.9165} & 0.8979          & \textbf{0.9870} & \textbf{0.9680} \\ \hline
\label{tab1}
\end{tabular}}
\end{table}

\begin{table}[!t]
\centering
\caption{Performance comparison with 13 medical segmentation models on ISIC 2018 dataset.}
\resizebox{0.75\linewidth}{!}{
\begin{tabular}{ccccc}
\hline
\textbf{Methods}     & \textbf{DSC}    & \textbf{SE}     & \textbf{SP}     & \textbf{ACC}    \\ \hline
U-Net \cite{ronneberger2015u}                & 0.8545          & 0.8800          & 0.9697          & 0.9404          \\
Att U-Net \cite{oktay2018attention}            & 0.8566          & 0.8674          & \textbf{0.9863} & 0.9376          \\
TransUNet \cite{chen2021transunet}            & 0.8499          & 0.8578          & 0.9653          & 0.9452          \\
MedT \cite{asadi2020multi}            & 0.8389          & 0.8252          & 0.9637          & 0.9358          \\
FAT-Net \cite{wu2022fat}              & 0.8903          & 0.9100          & 0.9699          & 0.9578          \\
TransNorm \cite{azad2022transnorm}            & 0.8951          & 0.8750          & 0.9790          & 0.9580          \\
MALUNet \cite{ruan2022malunet}              & 0.8931          & 0.8890          & 0.9725          & 0.9548          \\
ATTENTION SWIN U-NET \cite{aghdam2023attention} & 0.8540          & 0.8057          & 0.9826          & 0.9480          \\
SCR-Net \cite{wu2021precise}              & 0.8886          & 0.8892          & 0.9714          & 0.9547          \\
MSA \cite{wu2023medical}                 & 0.8829          & 0.9199          & 0.9745          & 0.9617          \\
HorNet \cite{rao2022hornet}               & 0.9020          & \textbf{0.9212} & 0.9645          & 0.9596          \\
C$^2$SDG \cite{gu2023cddsa}                & 0.8806          & 0.8970          & 0.9643          & 0.9506          \\
META-Unet \cite{wu2023meta}            & 0.8899          & 0.8909          & 0.9716          & 0.9552          \\
\textbf{MHA-UNet(Our)}    & \textbf{0.9074} & 0.9149          & 0.9649          & \textbf{0.9680} \\ \hline
\label{tab2}
\end{tabular}}
\end{table}

Table \ref{tab3} and \ref{tab4} show the results of experiments with 8 state-of-the-art medical segmentation models externally validated on PH$^2$ and our clinical dataset. We can conclude from the table that our generalization performance has a more pronounced effect. On both data, the DSC values of our model are 4.03$\%$ and 1.36$\%$ higher, respectively, when compared to the traditional high-order interaction model with a single order (HorNet). In the generalization experiments our model also has a strong explanatory segmentation process and the most accurate segmentation results.

\begin{figure}[!t]
\centering
\includegraphics[width=0.6\linewidth]{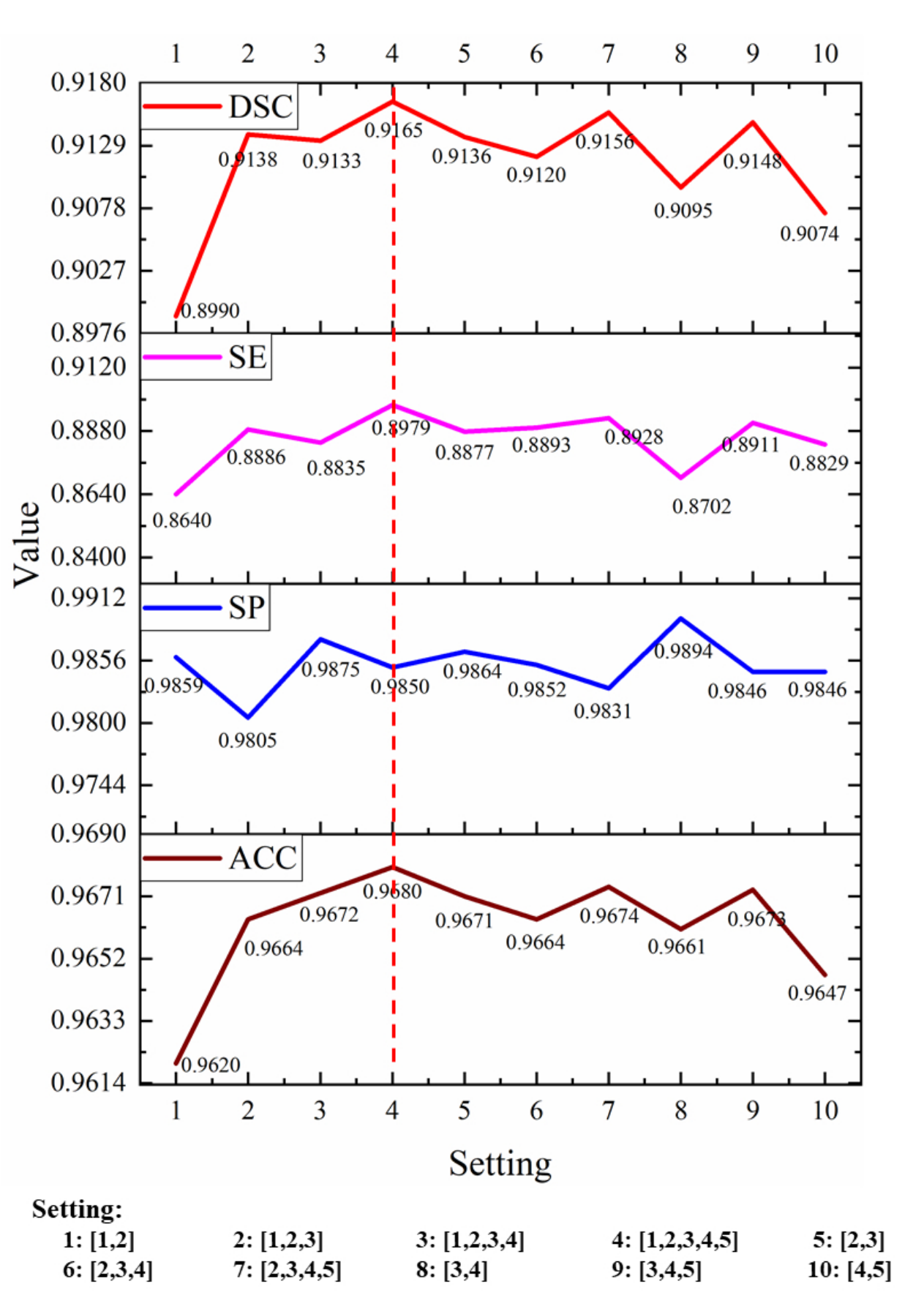}
\caption{Ablation experiments on the fusion of different attention interaction orders in MHAblock.}
\label{fig06}
\end{figure}

\subsection{Ablation study}
To further validate the effectiveness of our proposed Multiple High-order Attention Interaction Block (MHAblock), we conducted ablation experiments. The specific setup of the experiments is shown in Table \ref{tab5}. We replace the high-order attention interaction module with a single order in the encoder and decoder of the proposed MHA-UNet model, respectively. For the choice of single order we use the optimal interaction order proposed by HorNet \cite{rao2022hornet}. As shown in Table \ref{tab5}, setting $A$ indicates replacing the MHAblock of the encoder with a single order interaction. Setting $B$ indicates replacing the MHAblock of the decoder with a single order interaction. Setting $A+B$ indicates replacing both the encoder and decoder MHAblock with a single order. The experimental results demonstrate that all evaluation metrics decrease when replacing either the encoder or the decoder with a single order interaction. In particular, the performance is worst in the case of simultaneous encoder and decoder replacement. This ablation experiment further demonstrates the effectiveness of the proposed Multiple High-order Attention Interaction Block (MHAblock).

\begin{table}[!t]
\centering
\caption{External validation performance comparison with 8 state-of-the-art models on the PH$^2$ dataset.}
\resizebox{0.75\linewidth}{!}{
\begin{tabular}{ccccc}
\hline
\textbf{Methods}     & \textbf{DSC}    & \textbf{SE}     & \textbf{SP}     & \textbf{ACC}    \\ \hline
TransNorm \cite{azad2022transnorm}            & 0.8952          & 0.9116          & 0.9328          & 0.9261          \\
MALUNet \cite{ruan2022malunet}              & 0.8865          & 0.8922          & 0.9425          & 0.9263          \\
ATTENTION SWIN U-NET \cite{aghdam2023attention} & 0.8850          & 0.8886          & 0.9363          & 0.9213          \\
SCR-Net \cite{wu2021precise}             & 0.8989          & 0.9114          & 0.9446          & 0.9339          \\
MSA \cite{wu2023medical}                  & 0.9096          & \textbf{0.9623} & 0.9382          & 0.9401          \\
HorNet \cite{rao2022hornet}               & 0.8894          & 0.9567          & 0.9073          & 0.9232          \\
C$^2$SDG \cite{gu2023cddsa}                 & 0.9030          & 0.9137          & 0.9476          & 0.9367          \\
META-Unet \cite{wu2023meta}            & 0.8998          & 0.9020          & 0.9510          & 0.9352          \\
\textbf{MHA-UNet(Our)}    & \textbf{0.9253} & 0.9539          & \textbf{0.9486} & \textbf{0.9503} \\ \hline
\label{tab3}
\end{tabular}}
\end{table}

\begin{table}[!t]
\centering
\caption{External validation performance comparisons with 8 state-of-the-art models on our clinical dataset.}
\resizebox{0.75\linewidth}{!}{
\begin{tabular}{ccccc}
\hline
\textbf{Methods}     & \textbf{DSC}    & \textbf{SE}     & \textbf{SP}     & \textbf{ACC}    \\ \hline
TransNorm \cite{azad2022transnorm}            & 0.8436          & 0.8015          & \textbf{0.9752} & 0.9406          \\
MALUNet \cite{ruan2022malunet}              & 0.8394          & 0.8605          & 0.9507          & 0.9321          \\
ATTENTION SWIN U-NET \cite{aghdam2023attention} & 0.8258          & 0.8160          & 0.9573          & 0.9291          \\
SCR-Net \cite{wu2021precise}              & 0.8446          & 0.8122          & 0.9678          & 0.9314          \\
MSA \cite{wu2023medical}                   & 0.8543          & 0.9162          & 0.9520          & 0.9436          \\
HorNet \cite{rao2022hornet}               & 0.8660          & \textbf{0.9626} & 0.9324          & 0.9386          \\
C$^2$SDG \cite{gu2023cddsa}                & 0.8542          & 0.8460          & 0.9650          & 0.9405          \\
META-Unet \cite{wu2023meta}            & 0.8569          & 0.8498          & 0.9643          & 0.9325          \\
\textbf{MHA-UNet(Our)}    & \textbf{0.8778} & 0.8874          & 0.9650          & \textbf{0.9490} \\ \hline
\label{tab4}
\end{tabular}}
\end{table}

\begin{table}[!t]
\centering
\caption{Ablation experiments on the effect of multiple orders and single orders on performance in MHA-UNet.}
\resizebox{0.55\linewidth}{!}{
\begin{tabular}{ccccc}
\hline
\textbf{Setup}     & \textbf{DSC}    & \textbf{SE}     & \textbf{SP}     & \textbf{ACC}    \\ \hline
MHA-UNet(baseline) & \textbf{0.9165} & 0.8979          & \textbf{0.9870} & \textbf{0.9680} \\
A                  & 0.9145          & \textbf{0.9086} & 0.9809          & 0.9667          \\
B                  & 0.9130          & 0.8899          & 0.9855          & 0.9668          \\
A+B                & 0.9113          & 0.8826          & 0.9867          & 0.9664          \\ \hline
\label{tab5}
\end{tabular}}
\end{table}

In order to further demonstrate that the fusion of different orders can effectively improve the performance of the model, we conducted ablation experiments. The specific settings of the experiment are shown in Fig.\ref{fig06}. We set up 10 different settings respectively. We found an important conclusion that the 3-order attention interaction operation has a relatively large impact. In setup 2, adding the 3-order interaction gives a fast performance improvement. Whereas, in setup 10, removing the 3-order interaction showed a significant decrease in performance. This finding is further explained in the explainable process as well. As shown in Fig.\ref{fig03}, the 3-order interaction is an extraction of the overall location of the lesion, so the 3-order interaction plays a key role. Moreover, in setup 4, the optimal performance was obtained by performing the setup that fused the [1,2,3,4,5] orders. This ablation experiment combined with explainable segmentation also reflects that different orders have different degrees of influence.

\section{Discussion and explainable analysis}
In contrast to natural images, skin lesions usually have more disturbances, including skin epidermal shedding, hair interference, low contrast, and blurred boundaries, among others. The traditional higher-order interaction mechanism is difficult to learn the lesion features better, which is due to the lack of attention to the focus location. In this paper, we propose a high-order attention interaction mechanism. We introduce squeeze attention into the high-order interaction mechanism, and also make squeeze attention to pay attention to features in multiple high-levels. Moreover, to the best of our knowledge, we are the first to explain the mechanism of different order interaction mechanisms with explainable skin lesion segmentation. The proposed Multiple High-order Attention Interaction Block (MHAblock) that combines the different roles of each order for deep feature extraction.

\begin{figure}[!t]
\centering
\includegraphics[width=\linewidth]{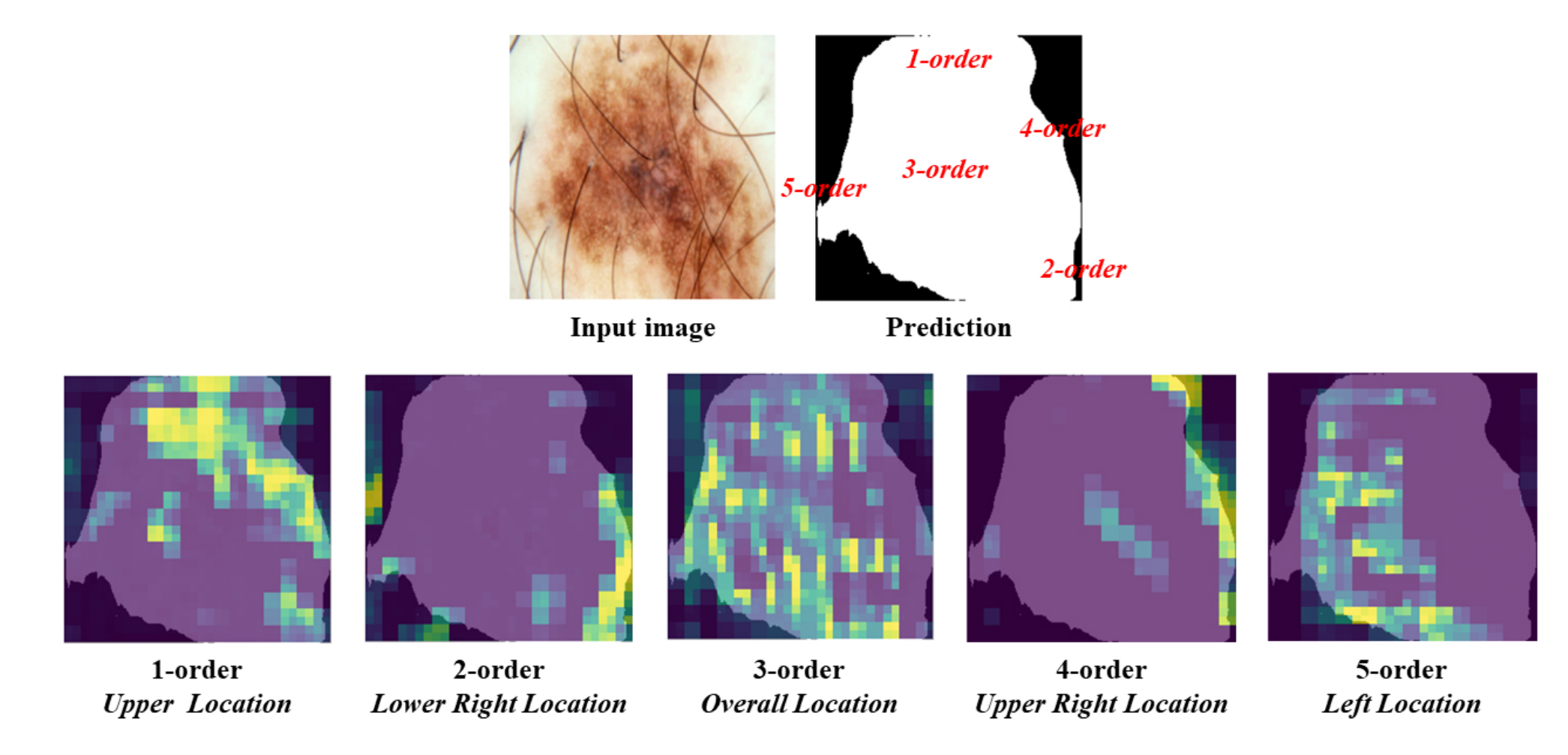}
\caption{Visualization plots of the characteristics of each attention interaction order in MHAblock.}
\label{fig07}
\end{figure}

\begin{figure}[!t]
\centering
\includegraphics[width=\linewidth]{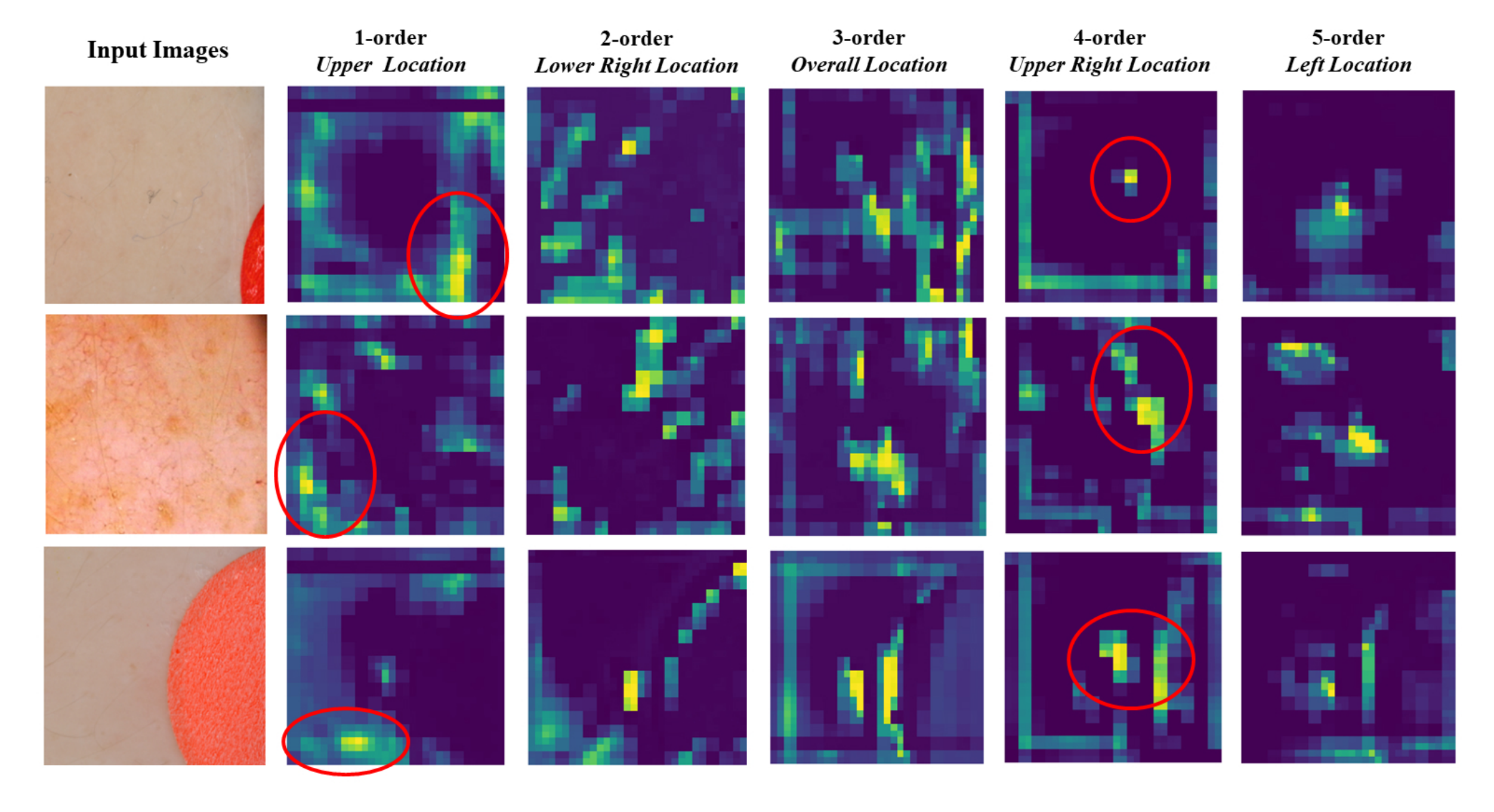}
\caption{MHAblock-based explainable analysis of negative samples without lesions.}
\label{fig08}
\end{figure}

\begin{table}[!t]
\centering
\caption{Experimental results for positive and negative categorization of multiple public datasets.}
\resizebox{0.45\linewidth}{!}{
\begin{tabular}{ccc}
\hline
\textbf{Datasets}     & \textbf{PDR} & \textbf{NDR} \\ \hline
PH$^2$                   & 81.0\%       & —            \\
Our clinical datasets & 79.5\%       & —            \\
NormalSkin            & —            & 78.5\%       \\
Kaggle95              & —            & 83.2\%       \\ \hline
\label{tab6}
\end{tabular}}
\end{table}

To further exemplify the skin lesion segmentation under interpretability, we visualize the results of different orders in the last MHAblock of the decoder. As shown in Fig.\ref{fig07}, the visualization graph we have performed some processing to display the prediction graph in combination with the order result graph, which helps to understand the role of each order interaction more clearly. As we can see from the figure, 1-order has a stronger sensitivity to the top feature information of the skin lesion, 2-order has a stronger perception of the lower right boundary information, 3-order is the most critical interaction order, which has the deepest learning of the overall feature information of the skin lesion, and 4-order has a stronger perception of the upper right boundary information. 5-order has a stronger perception of the left boundary information. We propose the Multiple High-order Attention Interaction Block (MHAblock) by combining five different orders of attention interactions. It is demonstrated in multiple experiments that our proposed MHA-UNet outperforms the current state-of-the-art models in skin lesion segmentation. In addition, the improvement in generalization ability is most exciting.

The positive case samples fully demonstrate the strong interpretability of our model. In addition, we intercepted negative samples without the presence of lesions in the ISIC2017 dataset as a test. As Fig.\ref{fig08} shows the visualization of the interpretability obtained from our test on negative samples. In particular, we circled the highlighted areas of 1-order and 4-order in red in Fig.\ref{fig08}. The 1-order and 4-order are performed as upper feature and upper right feature, respectively. And in the explainable analysis of the negative samples, we clearly see the wrong highlighted regions. This indicates the inability of the model to determine the location of the lesion in the inference process for the negative sample. In contrast to the positive samples, none of the visualized interpretations of the negative samples are compliant, with spurious highlighted regions. The reliability of our interpretable method can be demonstrated from both positive and negative samples.

Moreover, according to this strong interpretability, we perform classification under almost no introduction of additional memory by means of the proposed Explainable Inference Classification Algorithm (EICA). The adopted weights are trained on the segmentation task performed only with 1250 positive samples (ISIC2017 training set). As shown in Table \ref{tab6} are the results of our experiments, the positive detection rate on PH$^2$ and our clinical dataset reaches about 80$\%$. While the negative detection rate on NormalSkin and Kaggle95 tests is 78.5$\%$ and 83.2$\%$, respectively. This result demonstrates the reliability of our interpretability through quantitative aspects.

Although our results and interpretable analyses are exciting, we have only studied them in skin lesion segmentation experiments. In the future, the use of multiple higher-order attentional interaction mechanisms for diagnostic studies of a wider range of diseases will be an important direction. In addition, we used a small sample size of external clinical data. Although we only used it as an external validation, more clinical data for experiments would be a more comprehensive and accurate representation of the performance of the model. In the future, the collection of clinical samples is also an important direction. In addition, our proposed explainable inference classification algorithm (EICA) has considerable space for improvement. In the future, the classification accuracy can be effectively improved by designing a more appropriate EICA.

\section{Conclusion}
In this study, we propose high-order attention interaction mechanisms that introduce squeeze attention to multiple high-level feature attention. In addition, to the best of our knowledge, we are the first to explain the mechanism of different order interaction mechanisms. Meanwhile, we propose the multiple high-order attention interaction block (MHAblock) by fusing the features of different orders. The MHAblock is introduced into the UNet architecture and the MHA-UNet model is proposed. In addition, our method has very high interpretability in the skin lesion segmentation task, and our strong interpretability is further demonstrated from both positive and negative case samples. Comparison experiments of MHA-UNet with 13 medical segmentation models and external validation experiments with 8 state-of-the-art models demonstrate the superiority of our proposed method, along with its strong interpretability. In the future, applying the proposed method to the diagnosis of more diseases and increasing clinical sample collection will be a key direction. In addition, the existence of a large space of improvement for the Explainable Inference Classification Algorithm (EICA) is also a very interesting direction.

\section*{Acknowledgments}
This work was supported partly by Medical-Industrial Intersection Program of University of Shanghai for Science and Technology (2023LXY-RUIJINO1Z).


\end{document}